\documentclass{article}
\usepackage{spconf,amsmath,graphicx}
\usepackage{caption}
\usepackage{subcaption}

\title{AdvMS: A Multi-source Multi-cost Defense Against Adversarial Attacks}
\name{Xiao Wang$^{\star}$ \qquad Siyue Wang$^{\dagger}$ \qquad Pin-Yu Chen$^{\ddagger}$ \qquad Xue Lin$^{\dagger}$ \qquad Peter Chin$^{\star}$}
\address{$^{\star}$Boston University, Boston, MA, USA \\ 
$^{\dagger}$Northeastern University, Boston, MA, USA \\ 
$^{\ddagger}$IBM Research, Yorktown Heights, NY, USA}
\begin{document}
%
\maketitle

\begin{abstract}
 
Designing effective defense against adversarial attacks is a crucial topic as deep neural networks have been proliferated rapidly in many security-critical domains such as malware detection and self-driving cars. Conventional defense methods, although shown to be promising, are largely limited by their single-source single-cost nature: The robustness promotion tends to plateau when the defenses are made increasingly stronger while the cost tends to amplify. In this paper, we study principles of designing multi-source and multi-cost schemes where defense performance is boosted from multiple defending components. Based on this motivation, we propose a multi-source and multi-cost defense scheme, Adversarially Trained Model Switching (AdvMS), that inherits advantages from two leading schemes: adversarial training and random model switching. We show that the multi-source nature of AdvMS mitigates the performance plateauing issue and the multi-cost nature enables improving robustness at a flexible and adjustable combination of costs over different factors which can better suit specific restrictions and needs in practice.\let\thefootnote\relax\footnotetext{This work is supported by the Air Force Research Laboratory FA8750-18-2-0058, National Science Foundation DMS 1737897, and National Institute of Health R21 EY028381-01.} 
\end{abstract}


\section{Introduction}
Recent studies on adversarial attacks \cite{szegedy2013intriguing} reveal a disquieting circumstance that well-trained deep neural nets with high standard accuracy can be easily misled into making specific wrong predictions chosen aforehand by the adversary through crafting specific perturbations onto the inputs. The perturbed input examples, called adversarial examples, usually can be generated easily with gradient-based attacking methods, such as Fast Gradient Sign Method (FGSM) \cite{Goodfellow2015explaining}, Projected Gradient Descent (PGD) \cite{madry2018towards}, Carlini Wagner (CW) attack \cite{carlini2017towards,athalye2018obfuscated} and Elastic-net Attack (EAD) \cite{chen2017ead}. Adversarial examples are also shown to exist in the physical world \cite{kurakin2016adversarial,xu2019adversarial} and evasive even when viewed at different perspectives and scales \cite{athalye2017synthesizing}. These phenomena prompt a great need of designing effective and efficient defending schemes that insure robust and trustable learning systems.

Unfortunately, despite a rich body of defending mechanisms proposed recently in the literature \cite{madry2017towards,s.2018stochastic,zhou2018breaking,wang2018defensive,liu2017towards,sengupta2018mtdeep,zhou2018breaking,grosse2017statistical,metzen2017detecting,das2017keeping, wang2018defending}, there is no cost-free defense that has been discovered yet. Even if we assume additional costs in the training phase are totally acceptable (as training can be conducted off-line), the improved robustness using the above methods is often achieved at a price, such as drop of test accuracy \cite{madry2017towards,wang2018defensive}, more memory consumption \cite{sengupta2018mtdeep,zhou2018breaking, wang2019protecting} or false input rejection \cite{grosse2017statistical, metzen2017detecting} in the inference phase. In fact, the performance of a defending scheme is usually correlated with its cost, providing a trade-off between adversarial robustness and other performance factors. For example, adversarial training \cite{madry2017towards}, which is considered one of the most popular defending scheme in the literature, improves adversarial robustness of deep neural nets by including adversarial examples (with correct labels) in the training phase. In this method, although one can strengthen the defending effectiveness by using adversarial examples with large distortions, it also leads to degraded classification accuracy on natural images. In fact, although the ultimate goal of defenders is to design effective defense with negligible harms on other factors, some researchers also point out the trade-off between adversarial robustness and its cost factor (such as test accuracy) may be an inevitable  nature of deep neural nets \cite{su2018robustness,tsipras2018robustness}.

However, conventional single-source (that the robustness stems from a single defending scheme) and single-cost (that the robustness is traded with a single cost factor) defenses are often unable to provide a desirable robustness-cost trade-off. On the one hand, the defense improvement tends to exhaust as cost increases, leading to a decreasing marginal improvement of robustness. On the other hand, as robustness is traded with a single cost factor, it often exceeds the tolerance range of that factor in practice, making the defense invalid. 

In this paper, we study the principles of designing multi-source multi-cost defenses to break the above limitations. We anticipate that an efficient complex defense can be made with defense schemes with different source-cost pairs. Along with this intuition, we propose Adversarially Trained Model Switching (AdvMS) where a variety of sub-models adversarially trained  from different initializations are grouped together with a randomization scheme that assigns inputs to one of the sub-model randomly in the inference time. The selected sub-model is called to be activated which is ever-switching. The proposed defense method leverages the advantages of both adversarial training and stochastic defense and has two defending sources: adversarially trained parameters and stochasticity, and two cost factors: test accuracy and memory consumption. Empirical results illustrate its boosted defending performance and improved robustness-accuracy and robustness-memory trade-offs compared with adversarial training and model switching respectively. Moreover, the proposed method allows a more flexible robustness-accuracy-memory trade-off which is adjustable considering specific system needs.

The rest of the paper is organized as follows: motivations and details of the proposed approach is presented in Section \ref{motivation_advms}. Experimental results and analysis are provided in Section \ref{sec_experiments}, followed by Section \ref{Sec:Conclusion} that concludes this work.

\section{Adversarially Trained Model Switching (AdvMS)}
\label{motivation_advms}

\subsection{Motivation and Principles of Multi-source Multi-cost Defense}
We first explain the terms source and cost of a defense scheme using adversarial training \cite{madry2018towards} as an example, as it is considered one of the most advanced one in the literature. Adversarial training improves the worst-case robustness of a neural network by incorporating adversarial examples in the training process. As a result, the adversarially trained weights learn to adapt inputs with adversarial perturbations. We thus term the weights of adversarially trained models the source of the defense, as they are the surface of performing defense efforts. The cost of adversarial training is the test accuracy of the model on natural examples: the defense can be made stronger by using larger adversarial perturbations for training, at the cost of worse test accuracy on clean examples. 

We call defense schemes in this fashion single-source single-cost defenses, and most existing defenses fall into this category. As they are implemented alone, the drawbacks are obvious. Since there is only one scheme boosting robustness, the benefit in robustness tends to plateau with increased defense strength. Another issue is that the cost on a particular factor can easily exceed the acceptable range, making the defense strategy infeasible. 

A natural solution to the above limitations is to design multi-source multi-cost defense schemes. Although many proposed defenses claim to be compatible with others, designing complex defense is not trivial. We anticipate the following principles of combining multiple defense components as a complex defense. First, components need to have different defense source. This ensures the defenses are not intertwined which may cause degraded performance. Second, components need to have different defense cost. This avoids trading robustness with a single cost factor which leads to decreased marginal defense improvement, and at the same time it allows a more flexible combination of costs over multiple factors.

\subsection{Using Multi-Source Multi-Cost Defense: AdvMS}
\label{motivation_multi}
%

 



Following the above guidelines, we propose a complex defense AdvMS by merging adversarial training and switching-based stochastic defense. The scheme of model switching \cite{zhou2018breaking} protects the model by using a group of parallel sub-models trained from different random initialization, where the active one that processes model inputs is ever-switching in the run time. It will not degrade test accuracy but increases the memory consumption in the testing time. It thus has different source (stochasticity) and cost (memory consumption) compared to adversarial training. 


Implementing AdvMS with $M$ sub-models contains 3 steps: \textit{step 1}, randomly initialize the weights of $M$ sub-models of the same architecture and train them individually with the same training settings ($\epsilon_{train}$), and training data; \textit{step 2}, save all $M$ adversarially trained models and construct a pool of parallel sub-models; \textit{step 3}, add the stochastic scheme that randomly activate one model from the pool in each round of inference.



\section{Experimental Results and Analyses}
\label{sec_experiments}

\subsection{Defense Against White-box and Adaptive Attacks}
In this experiment, our proposed AdvMS is compared with its building blocks: adversarial training \cite{madry2017towards} and random model switching \cite{sengupta2018mtdeep,zhou2018breaking}. The purpose of this experiment is to show that the proposed complex defense outperforms each of its defending component performed alone.

\begin{table}[h]
\caption{Base Model Architectures for MNIST and CIFAR-10.}
\label{table_model_architecture}
\begin{center}
\scalebox{0.9}{
\begin{tabular}{|c|c|c|}
    \hline
     & MNIST  &  CIFAR-10\\
    \hline
     Conv layer & 32 $\times$ (3,3)  & 64 $\times$ (3,3) \\
     Conv layer & 32 $\times$ (3,3) & 64 $\times$ (3,3) \\
     Pooling layer & pool size (2,2) & pool size (2,2) \\
     Conv layer & 64 $\times$ (3,3) & 128 $\times$ (3,3) \\
     Conv layer & 64 $\times$ (3,3) & 128 $\times$ (3,3) \\
     Pooling layer & pool size (2,2) & pool size (2,2) \\
     Fully connected 1 & 200 units & 256 units \\
     Fully connected 2 & 200 units & 256 units \\
     Output layer & 10 units & 10 units \\
     \hline
\end{tabular}}
\end{center}
\end{table}

\begin{figure}[h]
    \centering
    \includegraphics[width=0.48\textwidth]{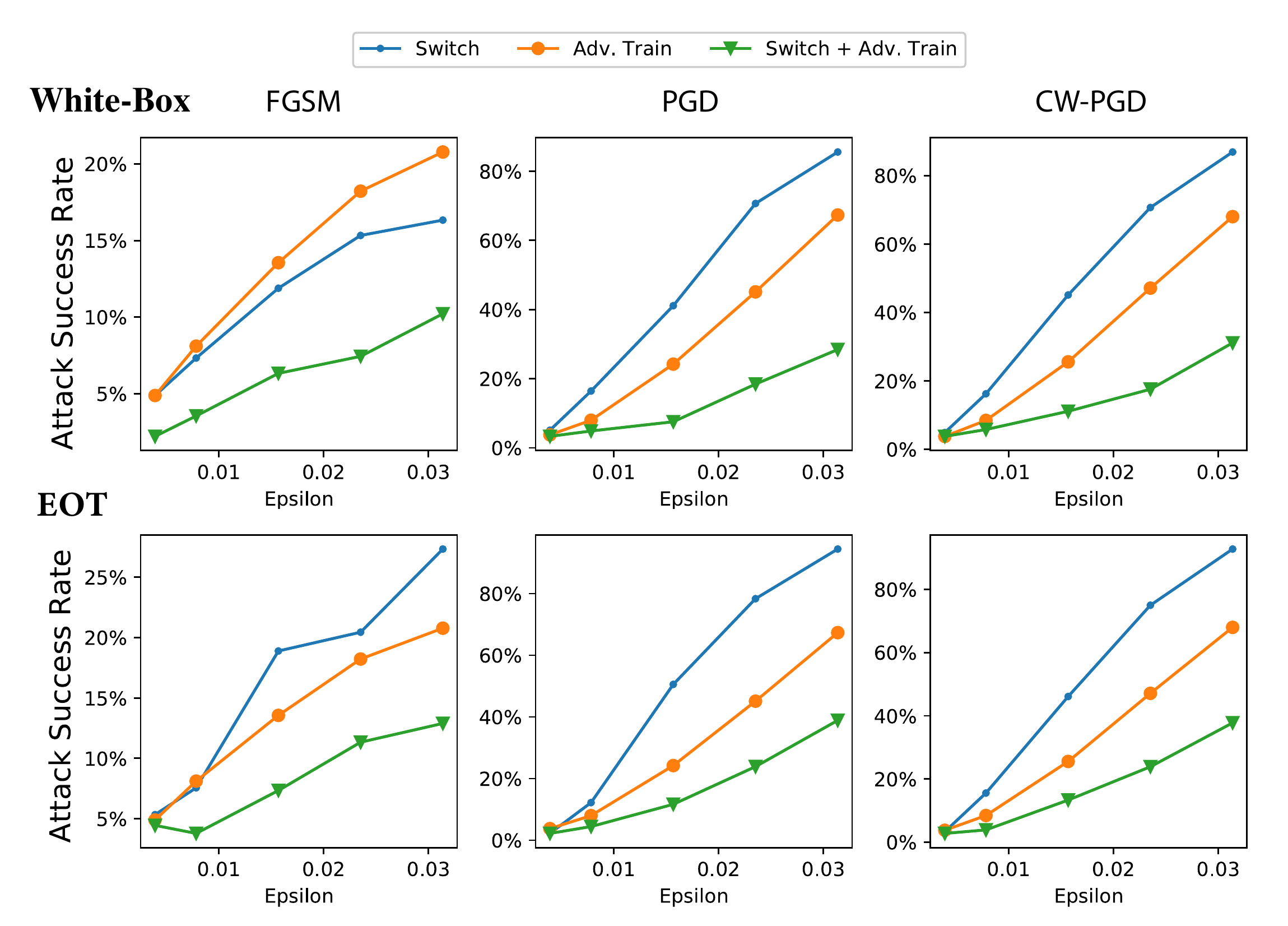}
    \caption{Attack success rate on CIFAR-10 dataset using  \textit{above:}(White-Box) FGSM, PGD, and CW-PGD attacks, \textit{below:} FGSM + EOT, PGD + EOT, and CW-PGD + EOT attacks}
    \vspace{-4mm}
    \label{fig:CIFAR_core_compare}
\end{figure}

\begin{figure}[h]
    \centering
    \includegraphics[width=0.48\textwidth]{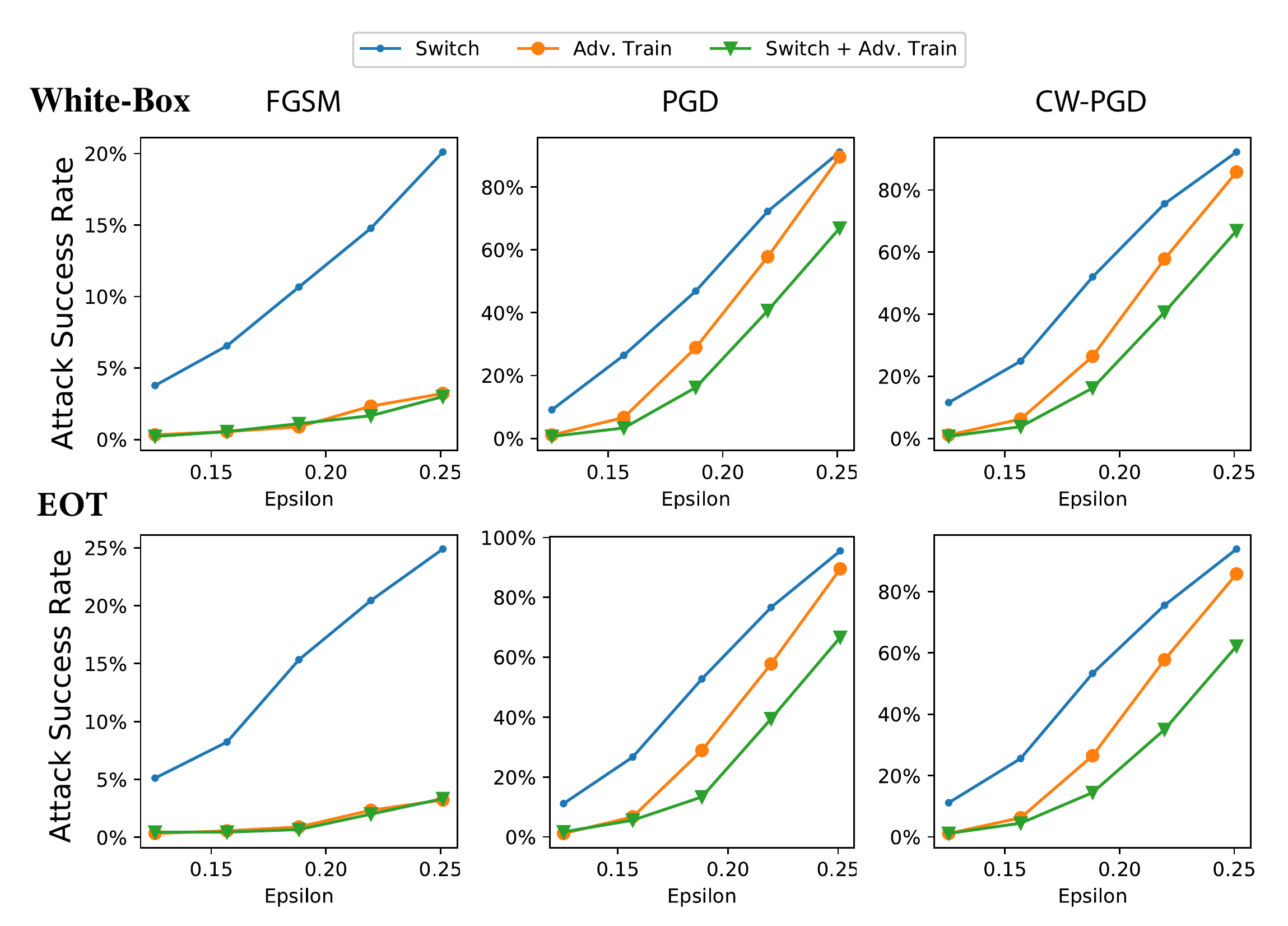}
    \caption{Attack success rate on MNIST dataset using  \textit{above:}  (White-Box) FGSM,   PGD,  and CW-PGD attacks, \textit{below:} FGSM + EOT,  PGD + EOT, and CW-PGD + EOT attacks.}
    \label{fig:MNIST_core_compare}
    \vspace{-2mm}
\end{figure}

For a fair comparison, all defenses are implemented on the same base model. Base model architectures for MNIST and CIFAR-10 are summarized in Table \ref{table_model_architecture}. The defending performance is measured by attack success rate (ASR) where lower ASRs indicate stronger defense. We perform three attacks: FGSM, PGD, and CW-PGD which are among the strongest baselines in the literature. All attacks are conducted in multiple strengths (controlled by the $L_\infty$ norm $\epsilon$ of the perturbation) and in both white-box setting, where we assume attackers know all information of the target model, and EOT (Expectation Over Transformation) \cite{athalye2018obfuscated,wang2019protecting} which is a counter-measure of attacks against randomized defense schemes by using the expectation of stochastic gradients. Fig.\ref{fig:CIFAR_core_compare} and Fig.\ref{fig:MNIST_core_compare} summarize the defending performance on CIFAR-10 and MNIST datasets respectively.

As we can observe, the proposed AdvMS achieves superior defense performance under all attack settings and performs even better under stronger attacks such as PGD and CW-PGD, where model switching and adversarial training are less effective given the same level of attack strength.




\subsection{Quantitative Analysis on Effects of Adversarial Training and Model Switching in AdvMS}
In this experiment, we perform a quantitative analysis by controlling each of the two strength factors of AdvMS: $\epsilon$, which is the $\ell_\infty$ distortion of adversarial examples used in training, and $M$ which is the number of models for switching. This experiment illustrates how enforcing one defending component contributes to boost the performance on the top of the other.

\begin{figure}[t]
    \centering
    \includegraphics[width=0.48\textwidth]{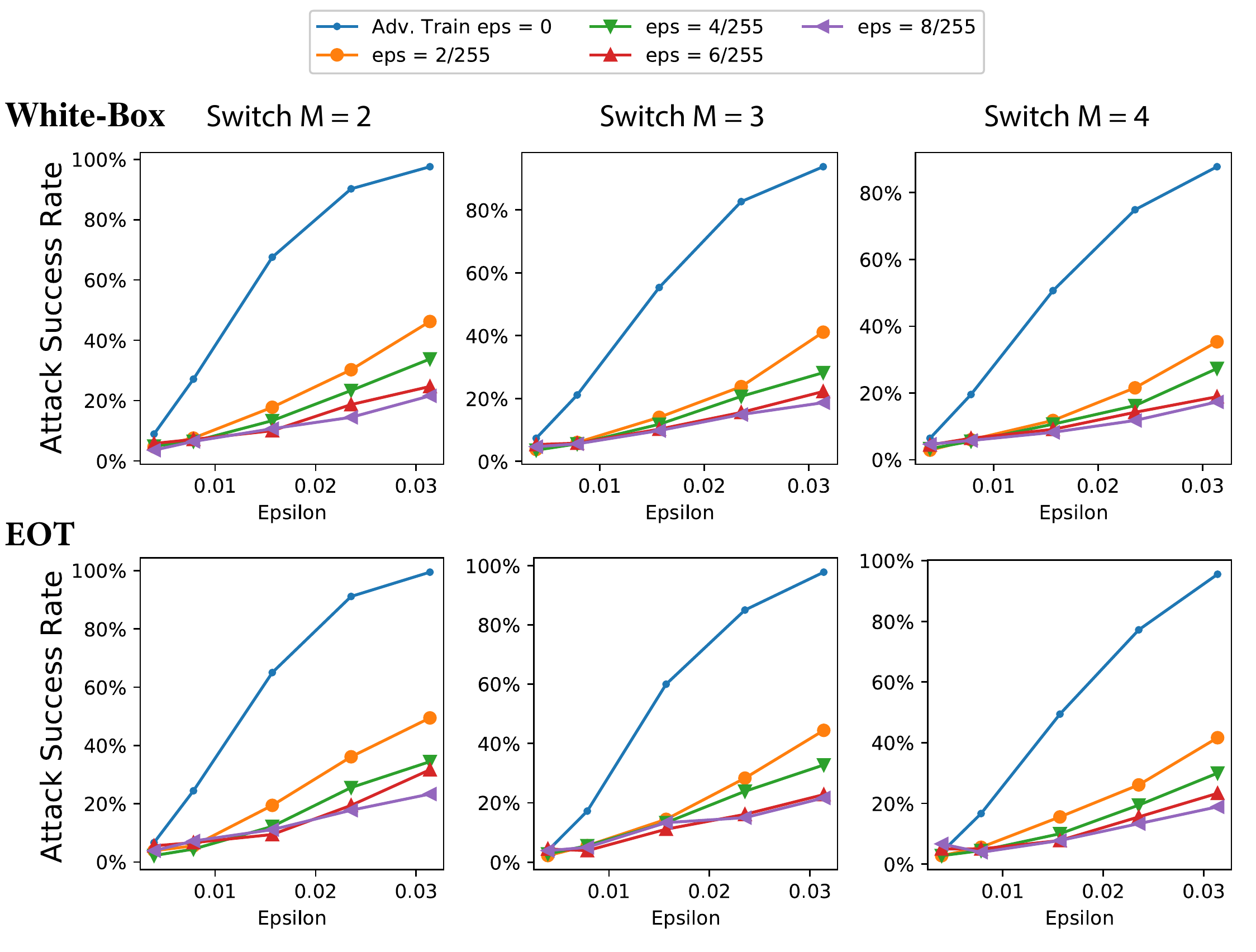}
    \caption{Attack success rate on CIFAR-10 dataset using different (white-box or EOT) PGD attack settings with fixed number of models $M$ in AdvMS}
    \label{fig:CIFAR_fix_M}
\end{figure}

In Fig.\ref{fig:CIFAR_fix_M}, we fix the number of models in AdvMS and compare the robustness performance obtained by changing the $\epsilon_{train}$ for adversarially trained models. In all cases we can observe the trend that, AdvMS trained with larger $\epsilon_{train}$  value shows stronger resiliency to adversarial attacks in both attack settings. However, it is also observed that although the gap between no adversarial training and adversarial training with $\epsilon_{train} = 2/255$ is large, the benefits gained by increasing $\epsilon_{train}$ tend to saturate with large $\epsilon_{train}$.

\begin{figure}[htbp]
    \centering
    \includegraphics[width=0.48\textwidth]{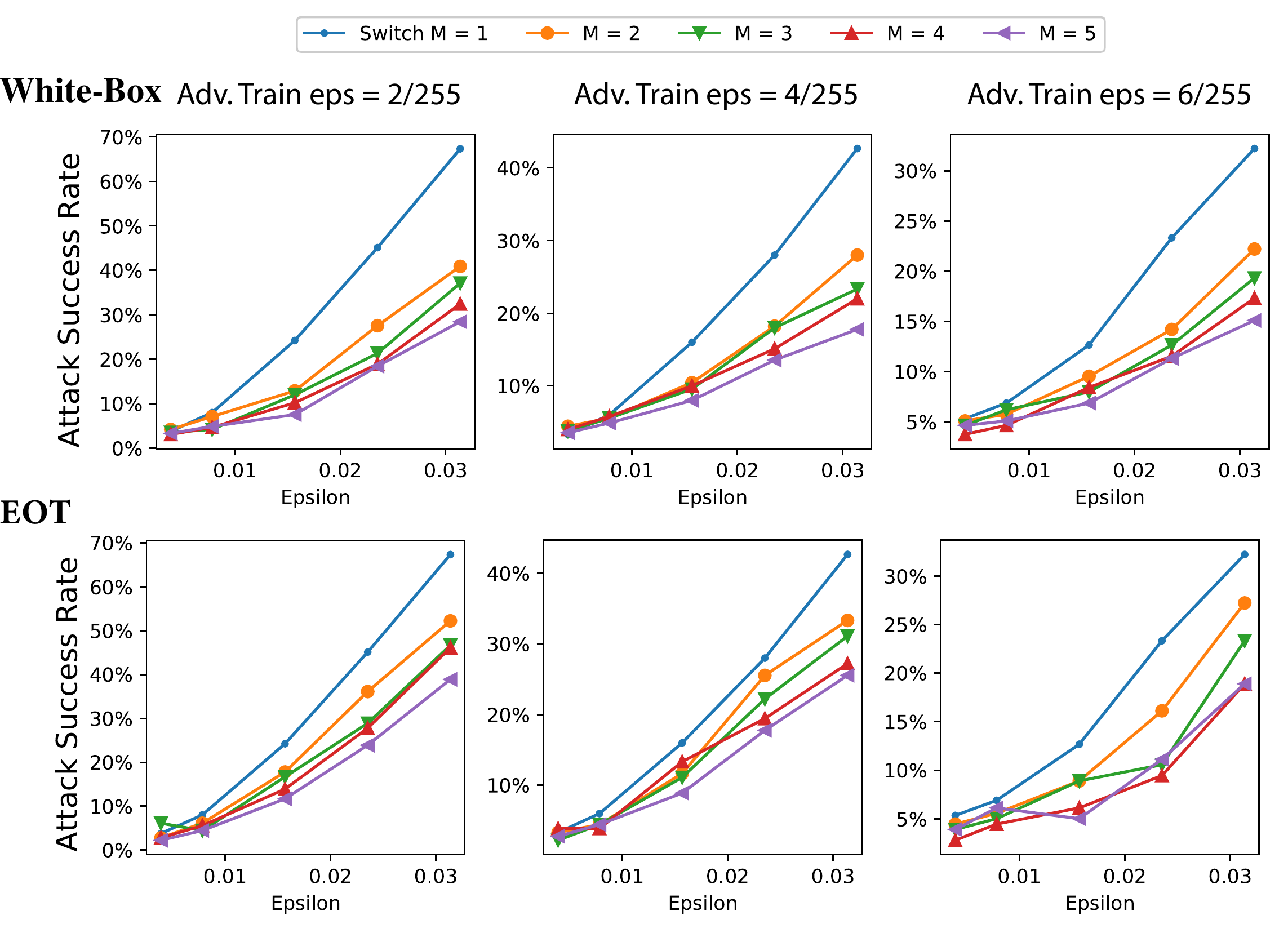}
    \caption{CIFAR-10 with fixed $\epsilon_{train}$ PGD Attack. Attack success rate on CIFAR-10 dataset using different (white-box or EOT) PGD attack settings with fixed adversarial training strength $\epsilon_{train}$ in AdvMS }
    \label{fig:CIFAR_fix_eps}
    \vspace{-3mm}
\end{figure}

We further perform quantitative analysis on changing the number of sub-models in AdvMS. Fig.\ref{fig:CIFAR_fix_eps} exhibits our findings that AdvMS with more adversarially trained models shows more resistance against adversarial attacks. While the gain in the defense rate when changing the deterministic model ($M = 1$) to a stochastic model with $M = 2$ is sufficiently large, the marginal gain in defense rate also tends to decease when $M$ is further increased.

These experiments illustrate the performance plateau issue as discussed earlier, and verify the importance of using multi-source defenses to break this limitation of single-source defenses.

\subsection{Quantitative Analysis on Robustness-Accuracy and Robustness-Memory Trade-offs of AdvMS}

\begin{figure}[t]
\centering
\begin{subfigure}{0.5\textwidth}
  \centering
  \includegraphics[width=0.95\linewidth]{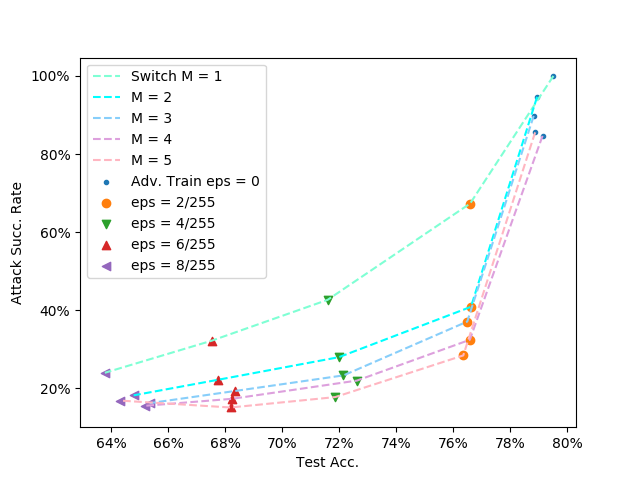}
  \caption{PGD}
  \label{fig:sub1}
\end{subfigure}%
\par
\begin{subfigure}{0.5\textwidth}
  \centering
  \includegraphics[width=0.95\linewidth]{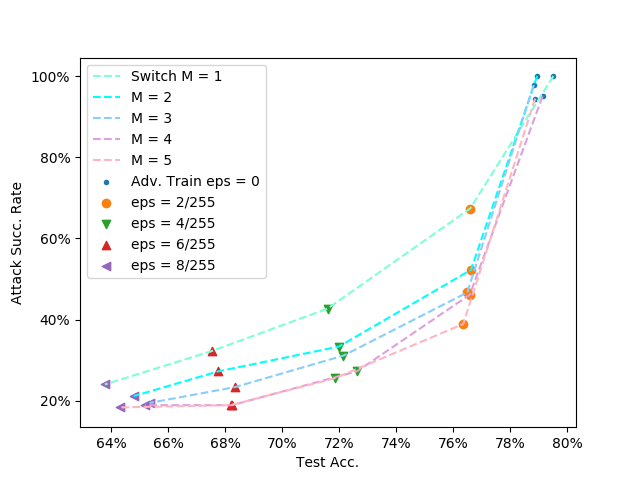}
  \caption{PGD + EOT}
  \label{fig:sub2}
\end{subfigure}
\caption{Robustness-Accuracy-Memory Trade-off on CIFAR-10 dataset. (a) Set $\epsilon_{attack} = 8/255$ for PGD attack. (b) Set $\epsilon_{attack} = 8/255$ for PGD attack + EOT (EOT: $n$=10).}
\vspace{-4mm}
\label{fig:PGD_vs_gm10}
\end{figure}

In Fig.\ref{fig:PGD_vs_gm10} (a) and Fig.\ref{fig:PGD_vs_gm10} (b), we plot defense effectiveness (quantified by ASRs) against test accuracy of defense models given by different combinations of Adversarial training strength $\epsilon$ and Number of Switching Models $M$ in white-box and EOT attack settings respectively. Here, points on the same curve are given by AdvMS models with the same $M$ and have the same amount of memory consumption. Points using the same marker type are given by AdvMS models with the same adversarial training distortion $\epsilon$ and have similar values of test accuracy. 

A defense scheme with ideal robustness-accuracy trade-off will result in a curve approaching to the bottom-right corner. As shown in Fig.\ref{fig:PGD_vs_gm10}, when increase $M$, the curve moves toward the bottom-right corner which indicates a better robustness-accuracy trade-off. For example, AdvMS with $M = 5$ can decrease ASR from 100\% to around 25\% with less than 3\% test accuracy drop while using adversarial training solely causes 15\% drop in test accuracy to achieve the same defense effect.

On the other hand, using large $\epsilon$ dramatically reduces the ASR, providing a better robustness-memory trade-off of the defense. It is also clear that the slope of curves become smaller on the left, which corresponds to the defense effectiveness plateau issue as discussed previously.

\section{Conclusion}
\label{Sec:Conclusion}
In this paper, we propose a novel multi-source multi-cost defense method: Adversarially Trained Model Switching (AdvMS) which can be implemented in the typical DNN model training pipeline and inherits advantages from both adversarial training and stochastic defense. Comparing to conventional single-source single-cost defenses, it is advantaged in preventing performance plateau issue and providing a more flexible trade-off among robustness, test accuracy and memory consumption. Experimental results demonstrate that it outperforms both adversarial training and model switching in defending both write-box and adaptive (EOT) attacks and provides improved robustness-accuracy and robustness-memory trade-offs compared with implementing the above two defenses alone respectively. 

\bibliographystyle{IEEEbib}
\bibliography{strings,refs}

\end{document}